\documentclass[10pt,twocolumn,letterpaper]{article}

\usepackage{cvpr}              % To produce the CAMERA-READY version
\usepackage{bm}
\usepackage{multirow}
\usepackage{amsmath} % if not already included
\newcommand{\norm}[1]{\left\lVert#1\right\rVert}

\definecolor{cvprblue}{rgb}{0.21,0.49,0.74}
\usepackage[pagebackref,breaklinks,colorlinks,allcolors=cvprblue]{hyperref}

\def\paperID{63} % *** Enter the Paper ID here
\def\confName{CVPR}
\def\confYear{2025}

\title{Paired and Unpaired Image to Image Translation using Generative Adversarial Networks}

\author{Gaurav Kumar \\
UC San Diego  \\
{\tt\small gkumar@ucsd.edu}
\and
Soham Satyadharma\\
UC San Diego\\
{\tt\small ssatyadh@ucsd.edu}
\and
Harpreet Singh\\
UC San Diego\\
{\tt\small h1singh@ucsd.edu}
}

\begin{document}
\maketitle
\begin{abstract}
Image-to-image translation is an active area of research in the field of Computer Vision, enabling the generation of new images with different styles, textures, or resolutions while preserving their characteristic properties. Recent architectures leverage Generative Adversarial Networks (GANs) to transform input images from one domain to another. In this work, we focus on the study of both paired and unpaired image translation across multiple image domains. For the paired task, we used a conditional GAN model, and for the unpaired task, we trained it with loss of cycle consistency. We experimented with different types of loss, multiple PatchGAN sizes, and model architectures. New quantitative metrics — precision, recall, and FID score — were used for quantitative analysis. In addition, a qualitative study of the results of different experiments was conducted.
\end{abstract}    
\section{Introduction}
\label{sec:intro}

Image-to-image translation refers to the class of problems where images are transformed from one domain (style/design) to another while preserving their characteristic features. 
% It is quite similar to the natural language translation task of converting text from one language to another, where the semantic meaning remains the same, but the representation (language) is different. 
The primary objective of such systems is to learn a mapping function between the input and output domains. Recent advances in Generative Adversarial Networks (GANs) \cite{goodfellow2014generative} have been successfully leveraged to address similar problems. 
% GANs are a zero sum game between two players: the generator and the discriminator, where the generator produces image of a certain domain in a quest to fool the discriminator into believing that the generated image is real. There can be multiple use cases of such a system.
GANs have been used to colorize black-and-white historical images \cite{Nazeri_2018} and to restore degraded images \cite{estrada2020deblurgan} using related techniques. 
% It can be also be used as a data augmenter. Gathering enough supervised training data for a machine learning problem is a challenging task, and people have been using GANs to generate training data.

In this paper, we address the problem of image translation using two approaches, wherein the model accepts an input image and generates a corresponding image from a different domain. First, we employ an image-level supervised learning technique, termed paired image-to-image translation, where paired samples are available for training. Second, we explore a domain-level supervised learning approach, referred to as unpaired image-to-image translation, where only sets of images from distinct domains are provided without explicit pairings. 
% This technique can be generalized easily to other tasks because the real bottleneck of any deep learning method is to gather enough labelled data for training. 
The contributions of this project are summarized as follows:

% For example, it is challenging to get a paired image dataset of famous paintings and their corresponding sketches. 
\setlist{nolistsep}
\begin{itemize}[noitemsep]
    \item Integration of the training processes for both paired and unpaired image translation
    \item Investigation of various loss metrics, including L1 loss, L2 loss, and their convex combination within the GAN loss framework, along with experimentation on different PatchGAN sizes and model architectures
    \item Evaluation of the results using quantitative metrics such as precision, recall, and inception FID score, supplemented by a qualitative analysis of the generated outputs to corroborate the quantitative findings
\end{itemize}

\section{Related Work}
\label{sec:related_work}

For the paired image-to-image translation task, Image Analogies \cite{hertzmann2001image} used non-parametric models learned using an autoregression algorithm. In contrast, our work utilizes parametric deep learning models, which extracts feature representations directly from images and generate higher-quality target domain images. Prior work has addressed specific image translation tasks, such as automatic image colorization \cite{zhang2016colorful, larsson2016learning}, semantic segmentation \cite{long2015fully}, and edge map detection \cite{xie2015holistically}, where deep CNNs were used to learn image feature representations. These methods typically employed unstructured losses (per-pixel regression or classification loss) that predict output pixels independently. In contrast, our approach leverages a structured GAN loss, which generates output image pixel values while preserving their dependency structure, resulting in improved predictions. Furthermore, our method can train a deep learning model with the same set of hyperparameters across multiple domain-transfer tasks.

For the unpaired image-to-image translation task, the Coupled Generative Network (CoGAN) \cite{liu2016coupled} learns a joint distribution of multi-domain images without using any labeled training image pairs, by employing a GAN for each domain and sharing a few weight layers among the GANs. However, inferring the joint distribution is challenging, as infinitely many joint distributions can arise from samples of marginal distributions. Liu et al. \cite{liu2017unsupervised} addressed this by enforcing a shared latent space assumption through the use of Variational Autoencoders, tying the last layers of the encoders and the initial layers of the decoders. Our work does not rely on such low-level similarity assumptions, thereby making our approach more flexible across different domains.

\section{Dataset}
\label{sec:dataset}

We used four open-source datasets for this project: CMP facades, maps, cityscapes, and horse to zebra. Except for the maps dataset, all other datasets contained $256\times256$ RGB images. The maps dataset consisted of $600\times 600$ RGB images, but we randomly cropped them to $256\times256$ during preprocessing. For the paired task, we used the facades, maps, and cityscapes datasets. For the unpaired task, we used the facades, maps, and horse to zebra datasets.

\textbf{CMP facades dataset} \cite{Tylecek13} consists of manually annotated facades collected from various sources across different cities worldwide, showcasing diverse architectural styles. It contains $606$ facade images paired with their corresponding architectural labels. 
    
\textbf{Maps dataset} \cite{isola2017image} comprises paired images of Google Maps and Google Earth regions. A total of $2194$ images were scraped from Google Maps around New York City.

\textbf{Cityscapes dataset} \cite{Cordts2016Cityscapes} features urban street scenes from $50$ European cities, captured during fair weather across different seasons. We used $3475$ images from this dataset. 
    
\textbf{Horse to zebra dataset} \cite{zhu2017unpaired} was curated by the authors of the paper using ImageNet \cite{russakovsky2015imagenet}, by searching the keywords {\textit{wild horse}} and {\textit{zebra}}. It contains $2661$ images in total. This is an unpaired dataset, as a horse does not have a corresponding zebra counterpart, and vice versa.

We did not create any explicit hand-engineered features. Since the images were passed through a CNN, the model learned relevant features directly from the data.

\subsection{Preprocessing}
We applied the following preprocessing steps to the images.
\setlist{nolistsep}
\begin{itemize} [noitemsep]
    \item \textit{Random flipping}: We generated a random number between $0$ and $100$ and flipped the image horizontally if the number was greater than $50$.
    \item \textit{Random jittering}: We resized the $256\times256$ images to $286\times286$ using nearest-neighbor extrapolation and then randomly cropped them back to $256\times256$.
    \item \textit{Normalization}: We normalized pixel values to the range \([-1, 1]\).
\end{itemize}

\section{Models}
\label{sec:models}

\subsection{Pix2pix Model}
In Pix2pix \cite{isola2017image}, conditional Generative Adversarial Networks (cGANs) are used. Traditional GANs take random noise as input to generate an output image belonging to the target distribution, whereas cGANs take an input sampled from distribution X and generate an output in another domain Y. Similar to GANs, cGANs consist of generator and discriminator networks. The discriminator aims to distinguish between real images from distribution Y and images generated by the generator from input x. The generator, on the other hand, tries to fool the discriminator into classifying generated images as real. The two networks are described in the following subsections.

\subsubsection{Generator}
The generator uses a deep CNN-based architecture, consisting of two parts: an encoder and a decoder. The encoder downsamples the input image while increasing its channel depth to form a latent representation. It comprises blocks of Conv-BatchNorm-ReLU layers. The decoder decompresses this latent representation using transpose convolution layers while decreasing the number of channels. The decoder consists of ConvTrans-BatchNorm-ReLU blocks.

The generator also incorporates skip-connections. Similar to the U-Net architecture \cite{ronneberger2015u}, the activations from the encoder are concatenated with the inputs to the corresponding decoder blocks. Dropout is applied to the first three layers of the decoder. Finally, a CNN layer reduces the number of output channels to match the input image channels.

\subsubsection{Discriminator}
A deep convolutional network is used as the discriminator, which takes an actual image sample from the target domain and a generated output, and computes a probability score indicating how likely they are to belong to different distributions. In this work, we use PatchGAN \cite{isola2017image}, also known as a Markovian discriminator. Instead of predicting a single probability score for the entire image, PatchGAN classifies different MxM patches as either real or generated, modeling image pixels as Markov random fields. It runs a deep convolutional network on the concatenated ground truth and generated images (via channel-wise concatenation).

\subsubsection{Loss Function}
Like standard GANs, conditional GANs use an adversarial loss function. Let G($\bm{x,z}$) represent the output from the generator G given input $\bm{x}$ and noise $\bm{z}$. Similarly, let D($\bm{x}$, G($\bm{x,z}$)) be the output of the discriminator D. The adversarial loss aims to improve the generator's ability to fool the discriminator and the discriminator's ability to distinguish between real and generated images in a minimax fashion. The adversarial loss is given by $L_{cGAN}$ in Equation 1. In addition, an L1 loss (Equation 2) is computed between the generated output and the ground truth image from the target domain to encourage the generator to produce outputs close to the real image, promoting sharper edges and reducing blurriness. The final objective, shown in Equation 3, combines the adversarial and L1 losses, with a regularization parameter $\lambda$ controlling the relative importance of the L1 loss.

\begin{equation}
\small
    \begin{split}
        L_{cGAN}(G, D) = & \mathbb{E}_{\bm{x,y}}[\log D(\bm{x,y})] + \\
                                &        \mathbb{E}_{\bm{x,z}}[\log (1 - D(\bm{x,G(x,z)}))]       
    \end{split}
\end{equation}
\begin{equation}
    \small
    L_{L1}(G) = \mathbb{E}_{\bm{x,y,z}}\big[  || \bm{y} - G(\bm{x,z}) ||_{1} \big]     
\end{equation}
\begin{equation}
    \small
    G^{*} = \arg \min_{G} \max_{D}  L_{cGAN}(G, D) + \lambda L_{L1}(G)
\end{equation}

\subsection{CycleGAN Model}

CycleGAN \cite{zhu2017unpaired} is used for unpaired image-to-image translation and builds upon the conditional GAN model described in Section 4.1. It consists of two conditional GANs: ($G_A, D_B$) and ($G_B, D_A$), where $G_A$ is a generator that takes an input image $\bm{I_A}$ from domain A and generates an image $\bm{I_A^{'}}$ that should belong to domain B. Similarly, $G_B$ is defined for the reverse mapping. The discriminator $D_B$ distinguishes between real images from domain B and generated images $\bm{I_A^{'}}$, and similarly for $D_A$.

A \textbf{Cycle Consistency Loss} is used to train the CycleGAN model. The generated image $\bm{I_A^{'}}$ is fed into the generator $G_B$, which attempts to reconstruct an image belonging to domain A. To enforce this consistency, an L1 loss is computed between $\bm{I_A}$ and $G_B(G_A(\bm{I_A}))$. Similarly, for an image $\bm{I_B}$ from domain B, an L1 loss is enforced between $\bm{I_B}$ and $G_A(G_B(\bm{I_B}))$. The total cycle consistency loss is the sum of these two L1 losses, as shown in Equation 4. For the two GANs ($G_A, D_B$) and ($G_B, D_A$), GAN losses are computed and added to the cycle consistency loss to form the full objective, given in Equation 6.

\setlength{\belowdisplayskip}{0pt} \setlength{\belowdisplayshortskip}{0pt}
\setlength{\abovedisplayskip}{0pt} \setlength{\abovedisplayshortskip}{0pt}
\begin{equation}
    \small
    \begin{split}
        L_{cyc}(G_A, G_B) & =  \mathbb{E}_{\bm{I_A\sim p_{data}(I_A)}}\big[  || G_B(G_A(\bm{I_A})) - \bm{I_A} ||_{1} \big] + \\
                        &  \mathbb{E}_{\bm{I_B\sim p_{data}(I_B)}}\big[  || G_A(G_B(\bm{I_B})) -  \bm{I_B}||_{1} \big]
    \end{split}
\end{equation}

\begin{equation}
    \small
    \begin{split}
        L_{cGAN}(G_A, D_B, A, B) = & \mathbb{E}_{\bm{I_B\sim p_{data}(I_B)}}\big[ \log D_B(\bm{I_B})\big] + \\
                        &  \mathbb{E}_{\bm{I_A\sim p_{data}(I_A)}}\big[ \log (1 - D_B(G_A(\bm{I_A})))  \big]
    \end{split}
\end{equation}

\begin{equation}
    \small
    \begin{split}
        L(G_A, G_B, D_A, D_B) = & L_{GAN}(G_A, D_B, A, B) + \\ 
                                    & L_{GAN}(G_B, D_A, B, A) + \lambda L_{cyc}(G_A, G_B)
    \end{split}
\end{equation}

\section{Experiments}
\label{sec:experiments}

We experimented by using different loss functions, varying Patch GAN sizes and removing skip connections from the U-Net \cite{ronneberger2015u} architecture to see how the results are affected. 
\subsection{Paired Task Experiments}
We trained the conditional GAN model using binary crossentropy loss as the GAN loss for the generator and also for the discriminator. We used the Adam optmizer for both networks with a learning rate of $2\times10^{-4}$ and $\beta_1$ as 0.5. One gradient descent step is performed for generator and one step for the discriminator. While performing gradient step for discriminator network, the loss objective is divided by 2 to slow down the rate of discriminator learning relative to generator. This is done to stabilize the GAN training. We tried different values of $\lambda$ but we got our best results with $\lambda=100$. We trained all models with a batch size of 16 for 150 epochs. For all experiments except the two Patch GAN experiments, we used a $70\times70$ Patch GAN.  

We conducted the experiments below for the paired task. 
\setlist{nolistsep}
\begin{enumerate}[noitemsep]
    \item L1 loss: L1 loss between generated and real images in addition to the GAN loss for the generator.
    \item L2 loss: L2 loss between generated and real images in addition to the GAN loss for the generator.
    \item $0.5 \times$ L1 + $0.5 \times$ L2: A combination of L1 and L2 loss to see which loss leads to better images. 
    \item Patch 16: Patch GAN of size 16x16.
    \item Patch 286: Patch GAN of size 286x286.
    \item Skip: UNet architecture without skip connections.
\end{enumerate}
\subsection{Unpaired Task Experiments}
Similar training settings as in paired task experiments are used to train unpaired task experiments.
\setlist{nolistsep}
\begin{enumerate}[noitemsep]
    \item L1 loss: L1 distance to compute cyclic loss
    \item L2 loss: Mean square error based cyclic loss
    \item $0.5 \times$ L1 + $0.5 \times$ L2: A convex combination of L1 and L2 cyclic loss
    % \item Patch 16: Patch GAN of size 16x16.
    % \item Patch 286: Patch GAN of size 286x286.
    % \item Skip: UNet architecture for the generator without skip connections.
\end{enumerate}

\begin{figure} [h!]
\begin{subfigure}[b]{0.23\textwidth}
    \centering
    \includegraphics[height=3cm, width=4cm]{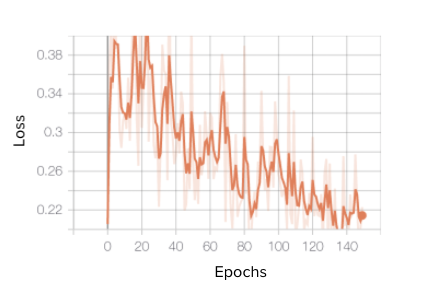}
    \caption{Generator L1 loss}
    \label{gen_l1_loss}
\end{subfigure}
\begin{subfigure}[b]{0.23\textwidth}
    \centering
    \includegraphics[height=3cm, width=4cm]{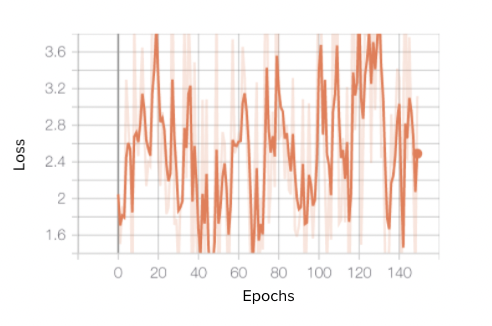}
    \caption{Generator GAN loss}
    \label{gen_gan_loss}
\end{subfigure}
\begin{subfigure}[b]{0.23\textwidth}
    \centering
    \includegraphics[height=3cm, width=4cm]{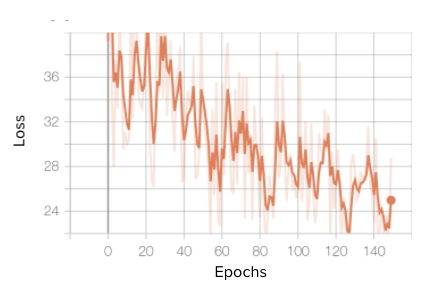}
    \caption{Generator total loss}
    \label{gen_tot_loss}
\end{subfigure}
\begin{subfigure}[b]{0.23\textwidth}
    \centering
    \includegraphics[height=3cm, width=4cm]{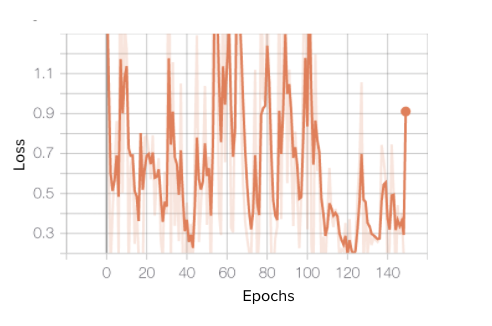}
    \caption{Discriminator loss}
    \label{disc_loss}
\end{subfigure}
\caption{Loss curves for the L1 loss experiment on the maps dataset with batch size=16 for 150 epochs}
\label{fig:loss}
\end{figure}

We have shown the training loss curves for the L1 loss experiment on the maps dataset in Figure \ref{fig:loss}. The generator and discriminator losses do not follow a set pattern as the generator tries to minimize the GAN loss, while the discriminator tries to minimize it. The generator L1 loss, which compares the generated and the real images, decreases with the number of epochs as the generator is generating better images. Similarly, the generator total loss, which is a combination of the generator GAN loss and the L1 loss also follows the decreasing trend of the L1 loss.

\section{Quantitative Results}
\label{sec:quant_results}

We evaluated our models using precision, recall, and Frechet Inception Distance (FID) scores on 256 generated images for each experiment.

The results are summarized in Table \ref{paired-table} and Table \ref{unpaired-table}.

\begin{table*}[!htbp]
\small
\centering
\begin{tabular}{|l|l|l|l|l|l|l|l|l|l|}
\hline
\multirow{2}{*}{\textbf{Experiment}}  & \multicolumn{3}{c|}{\textbf{Cityscapes}} & \multicolumn{3}{c|}{\textbf{Maps}} & \multicolumn{3}{c|}{\textbf{Facades}}    \\ \cline{2-10}
     & \textbf{Precision}  & \textbf{Recall} & \textbf{FID}      & \textbf{Precision} & \textbf{Recall} & \textbf{FID}      & \textbf{Precision} & \textbf{Recall} & \textbf{FID}      \\ \hline
L1 Loss         & 0.52       & 0.32   & 100.52 & 0.43      & 0.32   & 152.13 & 0.76      & 0.39   & 110.68  \\ \hline
L2 Loss         & 0.59       & 0.30   & 115.31 & 0.35      & 0.15   & 195.11 & 0.64      & 0.26   & 118.84 \\ \hline
0.5$\times$L1 + 0.5$\times$L2 & 0.61   & 0.19   & 108.70   & 0.42      & 0.05   & 205.14 & 0.71      & 0.39   & 116.89 \\ \hline
PatchGAN 16     & 0.19       & 0.23   & 147.51 & 0.33      & 0.06   & 211.69 & 0.76      & 0.29   & 116.50 \\ \hline
PatchGAN 286    & 0.48       & 0.19   & 154.22 & 0.31      & 0.15   & 198.18 & 0.73      & 0.21   & 124.90 \\ \hline
Skip            & 0.03       & 0.01   & 340.10 & 0.23      & 0.01   & 275.91 & 0.13      & 0.01   & 275.91 \\ \hline
\end{tabular}
\caption{\label{paired-table}\small Results on paired image-to-image translation tasks. All experiments were run with batch size 16 for 150 epochs. The L1 loss experiment shows the best overall performance.}
\label{tab:my-table}
\end{table*}

\subsection{Precision and Recall for GANs}
We computed precision and recall for GANs following the method proposed by Kynkäänniemi et al. \cite{kynkaanniemi2019improved}. The InceptionV3 network \cite{szegedy2016rethinking} was used to extract high-dimensional feature embeddings from the generated and real images. Let $\bm\Phi_g$ and $\bm\Phi_r$ denote the feature sets for generated and real images, respectively. 

For each $\bm\Phi \in \{\bm\Phi_g, \bm\Phi_r\}$, a hypersphere is defined around each feature vector with radius being distance to its $k^{th}$ nearest neighbor in $\bm\Phi$, using Euclidean distance. To determine if vector $\bm{\phi}$ lies within the hypersphere, we use:

% \begin{equation}
%     f(\bm{\phi}, \bm{\Phi})= 
% \begin{cases}
%     1, \quad \text{if} \quad \norm{\bm{\phi} - \bm{\phi^{'}}} \leq \norm{NN_k(\bm{\phi^{'}}, \bm\Phi) - \bm{\phi^{'}}}, \quad \forall \bm\phi^{'} \in \bm\Phi \\
%     0, \quad \text{otherwise}
% \end{cases}
% \end{equation}

\begin{equation}
f(\bm{\phi}, \bm{\Phi})= 
\begin{cases}
    1, & \text{if} \quad 
    \begin{aligned}[t]
        \norm{\bm{\phi} - \bm{\phi}^{'}} &\leq \norm{NN_k(\bm{\phi}^{\prime}, \bm{\Phi}) - \bm{\phi}^{\prime}}, \\
        &\forall\, \bm{\phi}^{\prime} \in \bm{\Phi}
    \end{aligned} \\
    0, & \text{otherwise}
\end{cases}
\end{equation}

Here, $NN_k(\bm{\phi^{'}}, \bm\Phi)$ denotes the $k^{th}$ nearest neighbor of $\bm\phi^{'}$ within $\bm\Phi$.  
If $\bm\phi \in \bm\Phi_g$ falls inside the hypersphere of any $\bm\phi^{'} \in \bm\Phi_r$, it is considered realistic.

Thus, precision and recall are defined as:

\begin{equation}
\begin{split}
\textrm{Precision} = \frac{1}{|\bm\Phi_g|} \sum_{\bm\phi \in \bm\Phi_g} f(\bm\phi, \bm\Phi_r)\\
\textrm{Recall} = \frac{1}{|\bm\Phi_r|} \sum_{\bm\phi \in \bm\Phi_r} f(\bm\phi, \bm\Phi_g)
\end{split}
\end{equation}

\subsection{Frechet Inception Distance (FID)}
The FID score \cite{heusel} measures the similarity between generated and real images by comparing their feature distributions. Images are passed through the pre-trained InceptionV3 network \cite{szegedy2016rethinking}, and features are extracted from the second-to-last layer. A lower FID score indicates higher fidelity and diversity in the generated samples.

Let $(\bm{\mu_r}, \bm{C_r})$ and $(\bm{\mu_g}, \bm{C_g})$ denote the mean vectors and covariance matrices of the real and generated feature embeddings, respectively. The FID score is calculated as:

\begin{equation}
    \textrm{FID} = \norm{\bm{\mu_r} - \bm{\mu_g}}^2 + \text{Tr}(\bm{C_r} + \bm{C_g} - 2(\bm{C_r}\bm{C_g})^{1/2})
\end{equation}

\begin{table}[!htbp]
\small
\begin{center}
\begin{tabular}{|l|r|r|r|}
\hline
\bf Experiment & \bf Maps & \bf Facades & \bf Horse2Zebra \\ \hline
L1 Cyclic Loss & 235.67 & 166.12 & 204.09 \\ \hline
L2 Cyclic Loss & 245.76 & 205.95 & 245.89 \\ \hline
0.5$\times$L1 + 0.5$\times$L2 & 298.45 & 187.34 & 211.78 \\ \hline
\end{tabular}
\end{center}
\caption{\label{unpaired-table}\small FID scores for unpaired image-to-image translation tasks. L1 cyclic loss achieves the lowest FID scores, aligning with our qualitative observations.}
\end{table}

\section{Qualitative Results}
\label{sec:qual_results}

\begin{figure} [h!]
\begin{subfigure}[b]{0.45\textwidth}
    \centering
    \includegraphics[height=2.7cm, width=7cm]{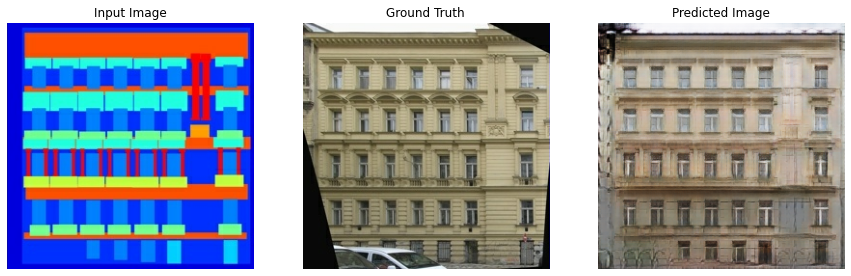}
    \caption{Facades Dataset Result}
    \label{facades_fig}
\end{subfigure}
\begin{subfigure}[b]{0.45\textwidth}
    \centering
    \includegraphics[height=2.7cm, width=7cm]{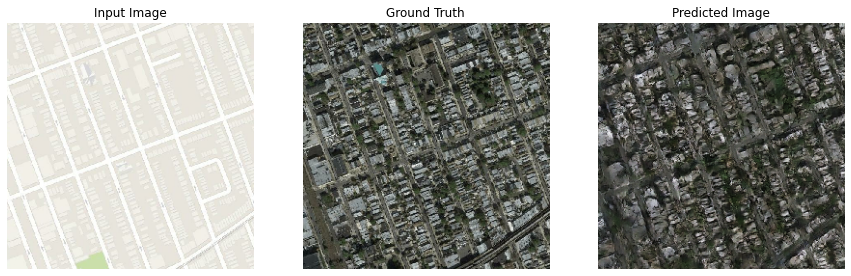}
    \caption{Maps Dataset Result}
    \label{maps_fig}
\end{subfigure}
\begin{subfigure}[b]{0.45\textwidth}
    \centering
    \includegraphics[height=2.7cm, width=7cm]{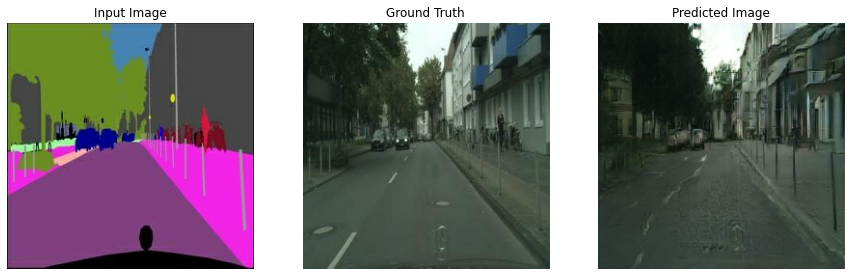}
    \caption{Cityscapes Dataset Result}
    \label{cityscapes_fig}
\end{subfigure}
\caption{From top to bottom: Input, ground truth, and generated images from paired L1 loss experiments on facades, maps, and cityscapes datasets}
\label{fig:paired_images}
\end{figure}

\begin{figure} [h!]
\begin{subfigure}[b]{0.45\textwidth}
    \centering
    \includegraphics[height=3.2cm, width=7cm]{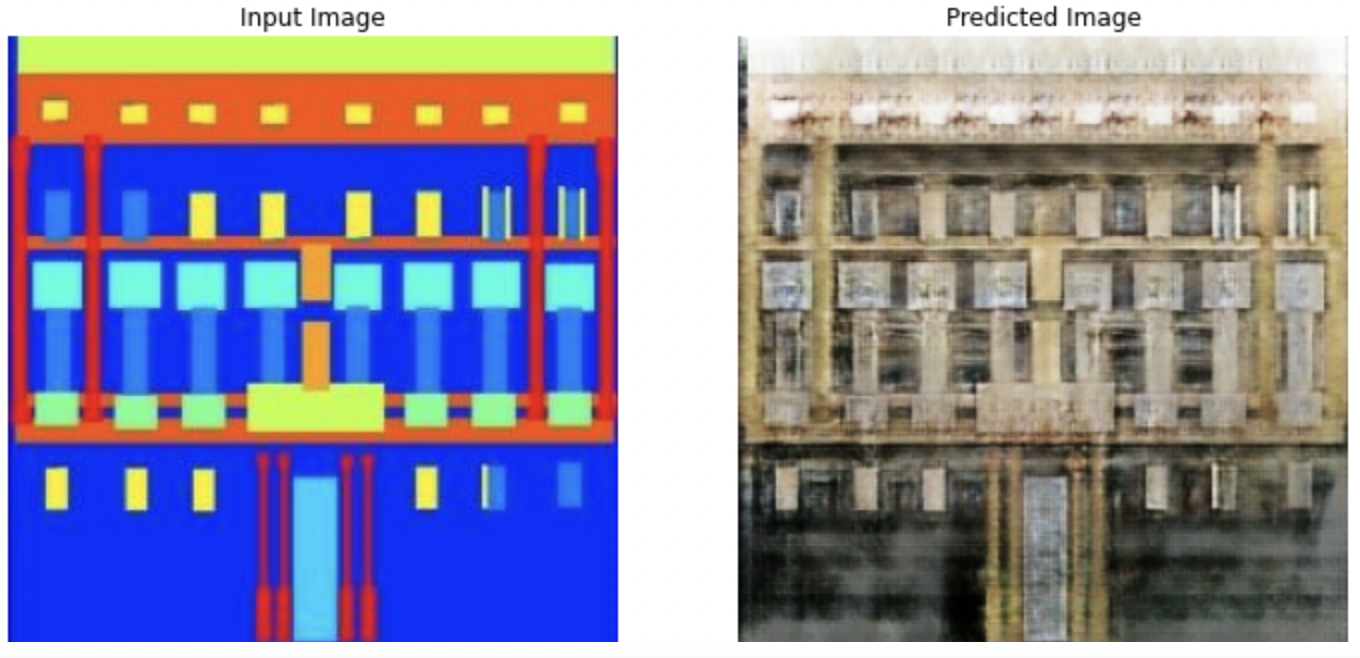}
    \caption{Facades Dataset Result}
    \label{facades_fig}
\end{subfigure}
\begin{subfigure}[b]{0.45\textwidth}
    \centering
    \includegraphics[height=3.2cm, width=7cm]{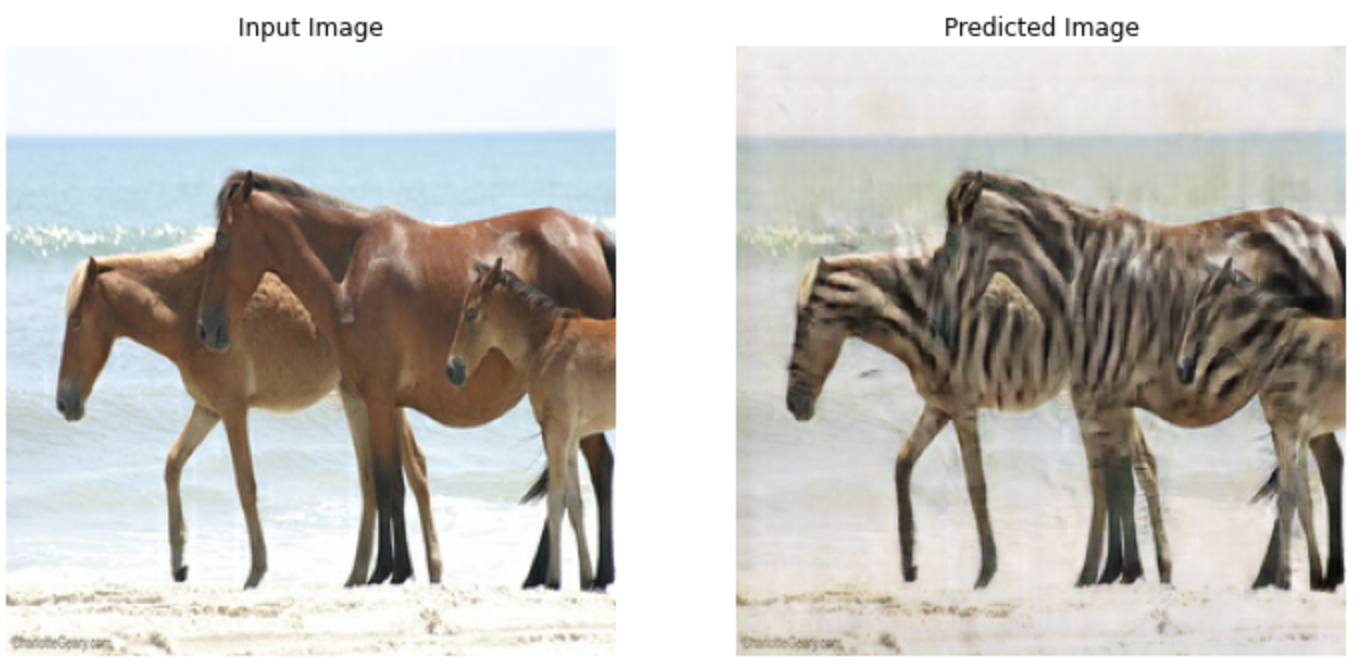}
    \caption{Maps Dataset Result}
    \label{maps_fig}
\end{subfigure}
\caption{From top to bottom: input and generated images from unpaired L1 cyclic loss experiments on facades and horse2zebra datasets.}
\label{fig:unpaired_images}
\end{figure}

\begin{figure} [h!]
\begin{subfigure}[b]{0.23\textwidth}
    \centering
    \includegraphics[height=2.5cm, width=3cm]{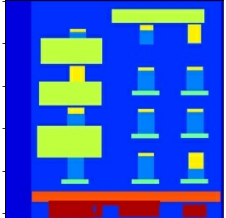}
    \caption{Input}
    \label{Input image}
\end{subfigure}
\begin{subfigure}[b]{0.23\textwidth}
    \centering
    \includegraphics[height=2.5cm, width=3cm]{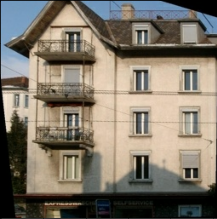}
    \caption{Target}
    \label{Target image}
\end{subfigure}
\begin{subfigure}[b]{0.23\textwidth}
    \centering
    \includegraphics[height=2.5cm, width=3cm]{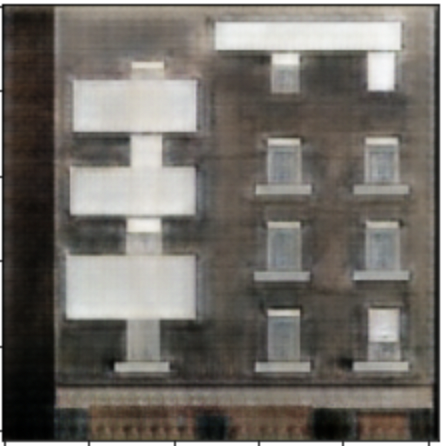}
    \caption{L1 cyclic loss}
    \label{l1 cyclic loss}
\end{subfigure}
\begin{subfigure}[b]{0.23\textwidth}
    \centering
    \includegraphics[height=2.5cm, width=3cm]{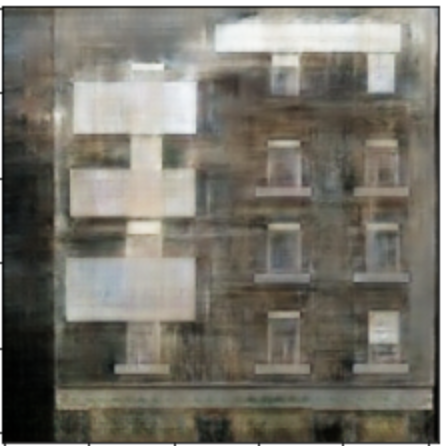}
    \caption{L2 cyclic loss}
    \label{l2 cyclic loss}
\end{subfigure}
\begin{subfigure}[b]{0.23\textwidth}
    \centering
    \includegraphics[height=2.5cm, width=3cm]{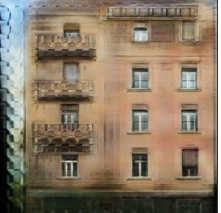}
    \caption{Patch 16}
    \label{pix2pix_16}
\end{subfigure}
\begin{subfigure}[b]{0.23\textwidth}
    \centering
    \includegraphics[height=2.5cm, width=3cm]{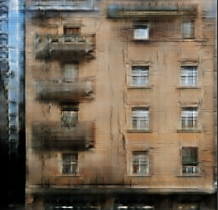}
    \caption{Patch 286}
    \label{pix2pix_16}
\end{subfigure}
\begin{subfigure}[b]{0.23\textwidth}
    \centering
    \includegraphics[height=2.5cm, width=3cm]{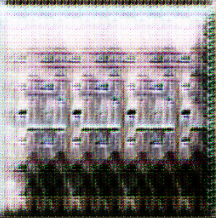}
    \caption{Skip}
    \label{pix2pix_16}
\end{subfigure}
\begin{subfigure}[b]{0.23\textwidth}
    \centering
    \includegraphics[height=2.5cm, width=3cm]{./Images/pix2pix_skip_facade}
    \caption{Pix2Pix L2}
    \label{pix2pix_l2_facade}
\end{subfigure}
\caption{Generation results using different methods. From left to right and top to bottom: Input, Ground Truth, L1 cyclic loss, L2 cyclic loss, Patch 16, Patch 286, Skip, and Pix2Pix L2 loss.}
\label{fig:other_results}
\end{figure}

This section qualitatively compares the Pix2Pix models learned across different experiments, evaluated on image ID 8 from the facades validation dataset. The results are presented in Figure \ref{fig:paired_images}, \ref{fig:unpaired_images}, \ref{fig:other_results}.

We observe that outputs generated by training the Pix2Pix model with L1 or L2 loss are of higher quality. These models benefit from image-level supervision, and the L1/L2 loss encourages the generator output distribution to closely match the target domain distribution. 

Furthermore, the image produced by the PatchGAN 16 experiment exhibits greater contrast compared to the one generated by the PatchGAN 286 experiment. This suggests that increasing the patch size makes the discriminator more tolerant to smoothness differences between real and generated images. In contrast, a smaller patch size enforces the generator to produce sharper edges and higher-contrast images, leading to improved visual quality.

Finally, we note that outputs from the paired task experiments generally achieve better quality than unpaired task, as the former benefits from direct ground truth supervision during training, whereas the latter does not.

\section{Conclusion}
\label{sec:conclusion}

In this work, we explored various generative methods for translating images from one domain to another. Our results demonstrate the effectiveness of paired image-to-image translation over unpaired translation across multiple datasets using the Pix2Pix GAN framework. We further show that incorporating L1 loss improves image quality consistently across different domains. In addition, we evaluated the models using quantitative metrics such as FID score, precision, and recall, which align with the qualitative improvements observed in the generated images.

\clearpage
{
    \small
    \bibliographystyle{ieeenat_fullname}
    \bibliography{main}

\begin{thebibliography}{19}
\providecommand{\natexlab}[1]{#1}
\providecommand{\url}[1]{\texttt{#1}}
\expandafter\ifx\csname urlstyle\endcsname\relax
  \providecommand{\doi}[1]{doi: #1}\else
  \providecommand{\doi}{doi: \begingroup \urlstyle{rm}\Url}\fi

\bibitem[Cordts et~al.(2016)Cordts, Omran, Ramos, Rehfeld, Enzweiler, Benenson, Franke, Roth, and Schiele]{Cordts2016Cityscapes}
Marius Cordts, Mohamed Omran, Sebastian Ramos, Timo Rehfeld, Markus Enzweiler, Rodrigo Benenson, Uwe Franke, Stefan Roth, and Bernt Schiele.
\newblock The cityscapes dataset for semantic urban scene understanding.
\newblock In \emph{Proc. of the IEEE Conference on Computer Vision and Pattern Recognition (CVPR)}, 2016.

\bibitem[Estrada et~al.(2020)Estrada, Lee, Dalgleish, Den~Ouden, Young, Smith, Desjardins, and Ouyang]{estrada2020deblurgan}
Dennis Estrada, Susanne Lee, Fraser Dalgleish, Casey Den~Ouden, Madison Young, Caitlin Smith, Joseph Desjardins, and Bing Ouyang.
\newblock Deblurgan-c: image restoration using gan and a correntropy based loss function in degraded visual environments.
\newblock In \emph{Big Data II: Learning, Analytics, and Applications}, page 1139507. International Society for Optics and Photonics, 2020.

\bibitem[Goodfellow et~al.(2014)Goodfellow, Pouget-Abadie, Mirza, Xu, Warde-Farley, Ozair, Courville, and Bengio]{goodfellow2014generative}
Ian~J. Goodfellow, Jean Pouget-Abadie, Mehdi Mirza, Bing Xu, David Warde-Farley, Sherjil Ozair, Aaron Courville, and Yoshua Bengio.
\newblock Generative adversarial networks, 2014.

\bibitem[Hertzmann et~al.(2001)Hertzmann, Jacobs, Oliver, Curless, and Salesin]{hertzmann2001image}
Aaron Hertzmann, Charles~E Jacobs, Nuria Oliver, Brian Curless, and David~H Salesin.
\newblock Image analogies.
\newblock In \emph{Proceedings of the 28th annual conference on Computer graphics and interactive techniques}, pages 327--340, 2001.

\bibitem[Heusel et~al.(2018)Heusel, Ramsauer, Unterthiner, Nessler, and Hochreiter]{heusel}
Martin Heusel, Hubert Ramsauer, Thomas Unterthiner, Bernhard Nessler, and Sepp Hochreiter.
\newblock Gans trained by a two time-scale update rule converge to a local nash equilibrium, 2018.

\bibitem[Isola et~al.(2017)Isola, Zhu, Zhou, and Efros]{isola2017image}
Phillip Isola, Jun-Yan Zhu, Tinghui Zhou, and Alexei~A Efros.
\newblock Image-to-image translation with conditional adversarial networks.
\newblock In \emph{Proceedings of the IEEE conference on computer vision and pattern recognition}, pages 1125--1134, 2017.

\bibitem[Kynk{\"a}{\"a}nniemi et~al.(2019)Kynk{\"a}{\"a}nniemi, Karras, Laine, Lehtinen, and Aila]{kynkaanniemi2019improved}
Tuomas Kynk{\"a}{\"a}nniemi, Tero Karras, Samuli Laine, Jaakko Lehtinen, and Timo Aila.
\newblock Improved precision and recall metric for assessing generative models.
\newblock \emph{arXiv preprint arXiv:1904.06991}, 2019.

\bibitem[Larsson et~al.(2016)Larsson, Maire, and Shakhnarovich]{larsson2016learning}
Gustav Larsson, Michael Maire, and Gregory Shakhnarovich.
\newblock Learning representations for automatic colorization.
\newblock In \emph{European conference on computer vision}, pages 577--593. Springer, 2016.

\bibitem[Liu and Tuzel(2016)]{liu2016coupled}
Ming-Yu Liu and Oncel Tuzel.
\newblock Coupled generative adversarial networks.
\newblock \emph{arXiv preprint arXiv:1606.07536}, 2016.

\bibitem[Liu et~al.(2017)Liu, Breuel, and Kautz]{liu2017unsupervised}
Ming-Yu Liu, Thomas Breuel, and Jan Kautz.
\newblock Unsupervised image-to-image translation networks.
\newblock \emph{arXiv preprint arXiv:1703.00848}, 2017.

\bibitem[Long et~al.(2015)Long, Shelhamer, and Darrell]{long2015fully}
Jonathan Long, Evan Shelhamer, and Trevor Darrell.
\newblock Fully convolutional networks for semantic segmentation.
\newblock In \emph{Proceedings of the IEEE conference on computer vision and pattern recognition}, pages 3431--3440, 2015.

\bibitem[Nazeri et~al.(2018)Nazeri, Ng, and Ebrahimi]{Nazeri_2018}
Kamyar Nazeri, Eric Ng, and Mehran Ebrahimi.
\newblock Image colorization using generative adversarial networks.
\newblock \emph{Lecture Notes in Computer Science}, page 85–94, 2018.

\bibitem[Ronneberger et~al.(2015)Ronneberger, Fischer, and Brox]{ronneberger2015u}
Olaf Ronneberger, Philipp Fischer, and Thomas Brox.
\newblock U-net: Convolutional networks for biomedical image segmentation.
\newblock In \emph{International Conference on Medical image computing and computer-assisted intervention}, pages 234--241. Springer, 2015.

\bibitem[Russakovsky et~al.(2015)Russakovsky, Deng, Su, Krause, Satheesh, Ma, Huang, Karpathy, Khosla, Bernstein, et~al.]{russakovsky2015imagenet}
Olga Russakovsky, Jia Deng, Hao Su, Jonathan Krause, Sanjeev Satheesh, Sean Ma, Zhiheng Huang, Andrej Karpathy, Aditya Khosla, Michael Bernstein, et~al.
\newblock Imagenet large scale visual recognition challenge.
\newblock \emph{International journal of computer vision}, 115\penalty0 (3):\penalty0 211--252, 2015.

\bibitem[Szegedy et~al.(2016)Szegedy, Vanhoucke, Ioffe, Shlens, and Wojna]{szegedy2016rethinking}
Christian Szegedy, Vincent Vanhoucke, Sergey Ioffe, Jon Shlens, and Zbigniew Wojna.
\newblock Rethinking the inception architecture for computer vision.
\newblock In \emph{Proceedings of the IEEE conference on computer vision and pattern recognition}, pages 2818--2826, 2016.

\bibitem[Tyle{\v c}ek and {\v S}{\' a}ra(2013)]{Tylecek13}
Radim Tyle{\v c}ek and Radim {\v S}{\' a}ra.
\newblock Spatial pattern templates for recognition of objects with regular structure.
\newblock In \emph{Proc. GCPR}, Saarbrucken, Germany, 2013.

\bibitem[Xie and Tu(2015)]{xie2015holistically}
Saining Xie and Zhuowen Tu.
\newblock Holistically-nested edge detection.
\newblock In \emph{Proceedings of the IEEE international conference on computer vision}, pages 1395--1403, 2015.

\bibitem[Zhang et~al.(2016)Zhang, Isola, and Efros]{zhang2016colorful}
Richard Zhang, Phillip Isola, and Alexei~A Efros.
\newblock Colorful image colorization.
\newblock In \emph{European conference on computer vision}, pages 649--666. Springer, 2016.

\bibitem[Zhu et~al.(2017)Zhu, Park, Isola, and Efros]{zhu2017unpaired}
Jun-Yan Zhu, Taesung Park, Phillip Isola, and Alexei~A Efros.
\newblock Unpaired image-to-image translation using cycle-consistent adversarial networks.
\newblock In \emph{Proceedings of the IEEE international conference on computer vision}, pages 2223--2232, 2017.

\end{thebibliography}
}

\clearpage
\appendix
\section{Appendix}
\label{sec:appendix}

This section showcases results from the experiments. 

% \begin{figure*} [t]
%     \includegraphics[height=2cm, width=\textwidth]{./Images/dataset_results}
%     \caption{From left: Input, ground truth, and generated images from paired L1 loss experiments on maps and cityscapes datasets; input and generated images from unpaired L1 cyclic loss experiments on facades and horse2zebra datasets.}
%     \label{maps}
% \end{figure*}

% \begin{figure}[!h]
% \centering
% \includegraphics[width=0.5\textwidth]{./Images/paired_facade}
% \caption{Input, ground truth, and generated images for paired L1 loss (maps, cityscapes) and unpaired L1 cyclic loss (facades, horse2zebra) experiments.}
% \label{maps}
% \end{figure}

% \begin{figure*} [t]
% \centering
% \includegraphics[height=2cm, width=\textwidth]{./Images/experiments.png}
% \caption{Images generated by different experiments on the facades dataset. From left to right: Input, Ground Truth, L1 cyclic loss, L2 cyclic loss, L1 loss, L2 loss, Patch 16, Patch 286, and Skip.}
% \label{input}
% \end{figure*}


\end{document}